# Prediction of Crash Injury Severity in Florida's Interstate-95

**B M Tazbiul Hassan Anik**
Graduate Research Assistant
Department of Civil, Environmental, and Construction Engineering
University of Central Florida, Orlando, FL 32816, USA.
Email: bmtazbiulhassan.anik@ucf.edu

**Md Mobasshir Rashid**
Graduate Research Assistant
Department of Civil, Environmental, and Construction Engineering
University of Central Florida, Orlando, FL 32816, USA.
Email: mdmobasshir.rashid@ucf.edu

**Md Jamil Ahsan**
Graduate Research Assistant
Department of Civil, Environmental, and Construction Engineering
University of Central Florida, Orlando, FL 32816, USA.
Email: mdjamil.ahsan@ucf.edu

\* **Corresponding author**

*Abstract*— Drivers can sustain serious injuries in traffic accidents. In this study, traffic crashes on Florida's Interstate-95 from 2016 to 2021 were gathered, and several classification methods were used to estimate the severity of driver injuries. In the feature selection method, logistic regression was applied. To compare model performances, various model assessment matrices such as accuracy, recall, and area under curve (AUC) were developed. The Adaboost algorithm outperformed the others in terms of recall and AUC. SHAP values were also generated to explain the classification model's results. This analytical study can be used to examine factors that contribute to the severity of driver injuries in crashes.

I. INTRODUCTION AND LITERATURE REVIEW

A. *Background and Problem*

One of the most significant social, economic, and health problems facing the world today is that of traffic accidents. Crash deaths are the leading cause of death for children and young adults aged 5 to 29, according to the World Health Organization's (WHO) global status report on road safety. Over 1.35 million people lose their lives due to automobile accidents each and every year (WHO, 2019). As Florida has consistently recorded high rates of crash rates over the years, the severity of injuries has been a pressing issue for both residents and authorities. To address this concern and raise awareness of the significance of road safety, it is essential to investigate the factors contributing to crash severity, most impacted demographics, and the ongoing efforts to mitigate this problem in Florida. This trend has far-reaching consequences, affecting not only the individuals and families directly affected by these events, but also the community as a whole in terms of public health, safety, and economic costs. To better understand the challenges posed by crash severity in Florida, it is necessary to execute a comprehensive analysis that illuminates the underlying factors, identify vulnerable populations, and propose potential interventions.

Numerous studies have been conducted in order to identify the risk variables that have a significant impact on the severity of an injury in the hopes of reducing the number of crash-related injuries. These elements that may have played a role can be classified under a number of different sections, including the environment, the road, the drivers, the vehicles, and the traffic. Injury severity prediction models are very significant safety tools that can reduce the impact of losses on both the economy and society. Other studies focused on assessing the crash injury risk of pedestrians during collision with vehicles (Jung et al., 2022; Rashid, 2022; Rashid et al., 2023; Rashid et al.). A significant amount of effort has been put into analyzing the factors connected to collisions and the severity of the injuries sustained in collisions. Researchers are now able to collect data and carry out a comprehensive study on traffic collisions, allowing them to predict the severity of crash injuries with high precision. This is made possible by the development of traffic detection systems over the past several years.

B. *Literature Review*

Crash severity prediction has gained significant attention in the field of transportation safety due to its potential to reduce traffic fatalities and severe injuries (Rabbani & Anik). Various data sources have been utilized for crash severity prediction, such as police-reported crash data, traffic flow data, and weather data (Li, Abdel-Aty, & Yuan, 2020). Recent studies have adopted several methodologies for crash severity prediction, including statistical, machine learning, and data mining techniques. Most statistical models show the mathematical relationship between accident characteristics and crash severity. Based on assumptions about uncertainty distribution and hypotheses tests, the statistical model can isolate the effects of different factors on collision severity (Chawla et al., 2002; Mansoor et al., 2020) Statistical methods, such as logistic regression (LR), multinomial logistic regression (MLR), and ordered probit models (OPM), have been widely used for crash severity prediction (Abdel-Aty & Radwan, 2000; AlMamlook et al., 2019). While statistical models focus on crash frequency and severity factors, machine learning algorithms, though effective in predicting outcomes (i.e., crash severity) in unexplored datasets, sometimes fall short in identifying specific factors due to their complex and often opaque nature (Ahsan et al., 2021; Ahsan & Siddique, 2021; Anik; Faden et al., 2023; Hasan et al., 2023a, 2023b; Kaiser & Ahsan, 2021; Raihan et al., 2023).

Machine learning techniques, including support vector machines (SVM), decision trees (DT), random forests (RF), and artificial neural networks (ANN), have been employed in recent studies (Li et al., 2021). These methods often provide higher accuracy compared to traditional statistical models, but require more complex computations. Deep learning techniques, such as convolutional neural networks (CNN) and recurrent neural networks (RNN), have been explored for crash severity prediction (Li, Abdel-Aty, et al., 2020). These methods can capture complex patterns and nonlinear relationships in the data, but require a large amount of training data and computational power.

Identifying critical factors influencing crash severity is vital for developing accurate models. Recent studies have explored various features, including roadway characteristics, traffic conditions, environmental factors, and driver behavior (Zeng et al., 2020). Variable selection techniques, such as stepwise regression and LASSO, have been employed to identify the most relevant features of traffic crashes (Zhu & Srinivasan, 2011). Additionally, intelligent transportation systems (ITS) have emerged as promising tools for real-time crash severity prediction. These systems integrate real-time data from multiple sources, such as road sensors and connected vehicles, to provide instant crash severity predictions (Rahim & Hassan, 2021). Real-time prediction has the potential to facilitate proactive safety interventions and mitigate crash severity. The real-time crash severity prediction has also been explored in Florida, integrating data from multiple sources such as road sensors and connected vehicles (Taheri et al., 2022).

*C. Proposed Work*

We investigated Florida District 5's I-95 crash statistics. Florida's east coast's major north-south interstate is I-95. From the Florida/Georgia state boundary to Miami International Airport, it covers much of the eastern mainland. There were 3,315 rear-end incidents, 1,839 side-impact crashes, 697 crashes with a fixed object, and 186 other crashes on Florida's I-95 between 2013 and 2015. Using the data from the collisions that took place in the area under study from 2016 to 2021, our objective was to make a prediction regarding the severity of the injuries sustained by drivers as a result of collisions. This study used a data-driven approach, with multiple classification algorithms to forecast the severity of driver injuries. For the purpose of prediction, we made use of several independent elements in our investigation. These included driver characteristics, vehicle characteristics, road characteristics, crash characteristics, temporal variables, and weather characteristics.

*D. ML Pipeline*

Figure 1 illustrates the ML pipeline used in this analytical study. After collecting data from crash data, we selected relevant features and did some preliminary exploration. The target variable is prediction of crash injury severity (serious/not serious). As most crashes lead to not serious injury, our dataset was highly imbalanced. We applied SMOTE algorithm to balance the dataset for two predictor classes (Anik et al., 2023; Raihan et al., 2023). Then we applied feature selection process via logistic regression method. After extracting important features, we applied several classification algorithms to predict crash injury severity and the model parameters were hyper-tuned for maximum performance. To achieve interpretability of our classification model, we obtained SHAP values.

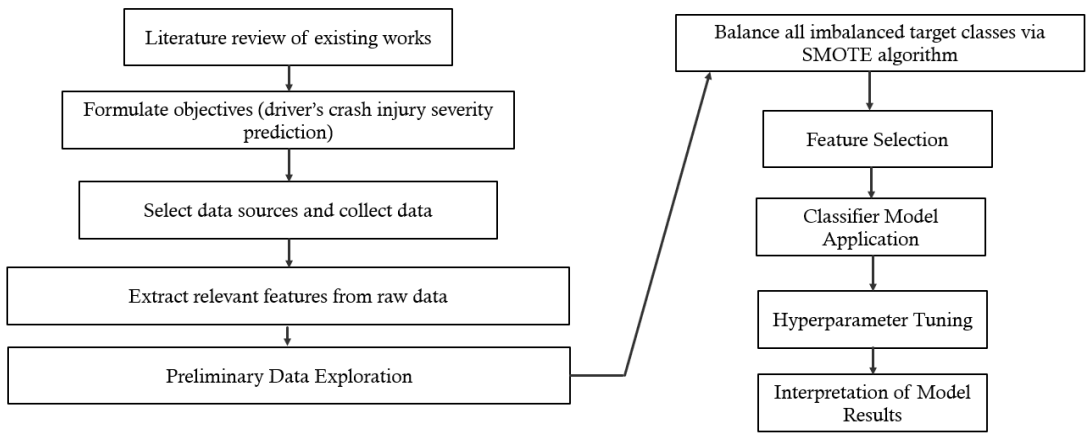

Fig. 1: ML pipeline used in the study.

## II. Data Processing

Signal Four Analytics (S4A) crash data was gathered for Florida's District 5 from 2016 to 2021. The components of the S4A database are organized into the following nine categories: Events, Vehicles, Drivers, Passengers, Non-Motorists, Violations, Pedestrian Typing, Bicyclist Typing, and Mapped Crashes. Each of these categories is further subdivided into subcategories. Our primary objective was to assess the severity of driver crash injuries, thus in order to accomplish this, we decided to employ event databases, vehicle databases, and driver databases. There were 4520 crashes with 18 features after integrating all datasets and filtering out missing information. The characteristics of the drivers, vehicles, roads, crashes, temporal patterns, and weather patterns that were present during the crashes were used to choose the features that are presented in Table I.

TABLE I. List of Features

| Type | Features | Description |
|---|---|---|
| Driver characteristics | Gender | Male or Female |
| | Sobriety Condition | Sober or Not Sober |
| | Age | Continuous |
| | Aggressive Driving/Speeding | Yes or No |
| Vehicle characteristics | Vehicle Type | Car, SUV, Heavy Vehicle |
| | Estimated speed (mph) | Continuous |
| | Vehicle Maneuver | Stopped, Straight, Lane Changing, Slowing, Turning, Other |
| Road characteristics | Road Surface Condition | Dry or Wet |
| | Road Alignment | Straight or Curve |
| Crash characteristics | Crash Type | Read End, Sideswipe, Other |
| | Crash Location | Intersection or Not-intersection |
| Temporal characteristics | Season | Winter, Summer, Spring, Fall |
| | Day of Week | Weekday or Weekend |
| | Time | Morning Peak, Morning Off-Peak, Day Off-Peak, Night Off-Peak, Afternoon Peak |
| Weather characteristics | Weather Condition | Clear, Cloudy, Rain |
| Area type | Urban or Rural | Rural or Urban |
| Traffic characteristics | Traffic Control Device | Controlled or Uncontrolled |
| | Light Condition | Daylight, Dark-lighted, Dark-not lighted |

### A. Variable Description

**This study's objective is to provide an estimate of the severity of injuries sustained in accidents, which is a binary variable with two possible values: serious injury and non-serious injury.** According to the final dataset, around 88% of injuries sustained in collisions were not serious, whereas only 12% were life-threatening. Figure 2 depicts the severity distribution of accident injuries in our investigation. The **SMOTE technique** was employed on the training data because the dataset was severely unbalanced. Random resampling such as SMOTE is used for rebalancing the minority class distribution in an uneven dataset.

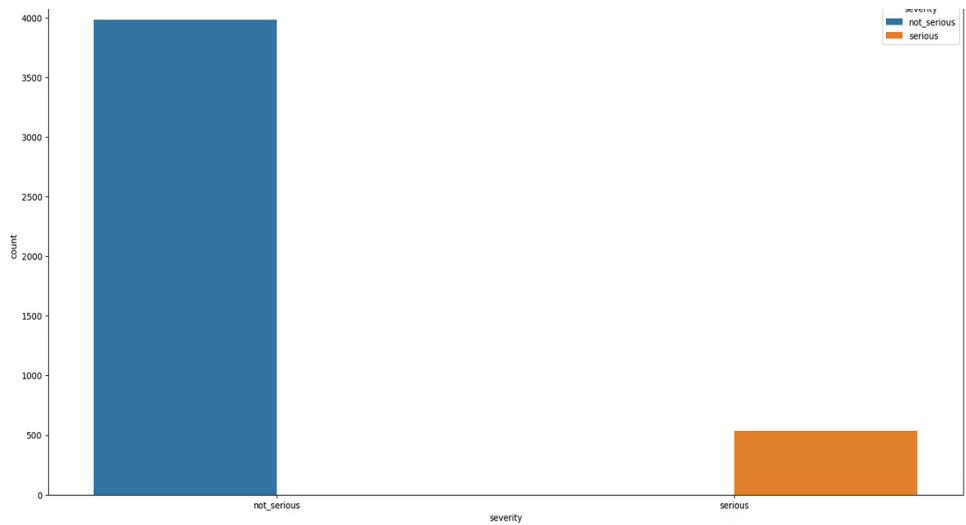

Fig. 2. Distribution of injury severity

*B. Feature Selection*

It's possible that a lot of the predictors are highly correlated with one another, or that the model has multicollinearity. In order to solve this problem, we employed a method called feature selection. This method reduced the amount of input variables by using only the most significant qualities and got rid of the data noise. Also, it is particularly beneficial to select features in order to reduce the total number of parameters, the amount of time spent training, the risk of overfitting as a result of enhanced generalization, and the impact of the curse of dimensionality. The logistic regression was used in other studies to understand the impact of associated factors on the dependent variable (Anik et al.; Rashid et al., 2018). Logistic regression was used to determine what features were the most significant with an interval of confidence of 90%. The most significant features generated from the logistic regression are shown in (Table II).

TABLE II. FEATURE SELECTION SCORES OF INPUT FEATURES

| Features | Coef. | Std. Err. | z | P>\|z\| | [0.025 | 0.975] |
| --- | --- | --- | --- | --- | --- | --- |
| driver_age_41_50 | 0.416514 | 0.193003 | 2.158072 | 0.030922 | 0.038235 | 0.794794 |
| driver_sobriety_condition_Sober | -1.61658 | 0.213813 | -7.56071 | 4.01E-14 | -2.03565 | -1.19751 |
| driver_age_more_60 | 0.56148 | 0.199769 | 2.81065 | 0.004944 | 0.16994 | 0.953019 |
| vehicle_type_Heavy_Vehicle | 0.68137 | 0.258692 | 2.633901 | 0.008441 | 0.174342 | 1.188397 |
| vehicle_type_SUV | 0.458464 | 0.198978 | 2.304097 | 0.021217 | 0.068475 | 0.848453 |
| vehicle_year_more_10 | 0.441557 | 0.144986 | 3.045511 | 0.002323 | 0.157389 | 0.725724 |
| crash_type_Rear_End | -0.31719 | 0.13907 | -2.28077 | 0.022562 | -0.58976 | -0.04461 |
| crash_type_Sideswipe | -0.41792 | 0.164028 | -2.54786 | 0.010839 | -0.73941 | -0.09643 |
| traffic_control_Uncontrolled | 0.971158 | 0.369798 | 2.626186 | 0.008635 | 0.246367 | 1.695948 |
| light_condition_Dark_Lighted | -0.63837 | 0.369417 | -1.72805 | 0.083979 | -1.36242 | 0.085672 |
| weather_condition_Clear | -0.49714 | 0.168453 | -2.95122 | 0.003165 | -0.8273 | -0.16698 |
| area_type_Rural | -0.34533 | 0.123733 | -2.79093 | 0.005256 | -0.58784 | -0.10282 |

### III. HYPERPARAMETER TUNING

Model parameters are learned from data in order to get the greatest fit, while hyper-parameters are set at the beginning of the training phase to ensure consistency. The use of search algorithms such as grid search and random search is necessary since finding the optimal values for hyper-parameters can be a time-consuming process. To adjust all hyperparameters of the multiple classification models utilized in this work, we employed Grid Search Cross-validation techniques. We used the logistic regression, decision tree classifier, random forest classification, SVM, adaptive boosting (Adaboost) classifier, and XGBoost classifier models. (Table III) displays the tuned hyperparameters of various classifiers.

TABLE III. TUNED HYPERPARAMETERS

| Model | Best Parameters |
|---|---|
| Logistic Regression | {'C': 0.0015, 'penalty': 'l2', 'solver': 'newton-cg'} |
| Decision Tree | {'criterion': 'entropy', 'max_depth': 23, 'min_samples_leaf': 1} |
| Random Forest | {'criterion': 'gini', 'max_depth': 19, 'max_features': 'sqrt', 'n_estimators': 50} |
| Support Vector Machine | {'C': 9, 'gamma': 'auto', 'kernel': 'rbf'} |
| Adaptive Boosting (Adaboost) | {'algorithm': 'SAMME.R', 'learning_rate': 0.8500000000000001, 'n_estimators': 25} |
| Extreme Gradient Boosting Decision Tree (XGBoost) | {'gamma': 1, 'learning_rate': 0.75, 'max_depth': 13, 'n_estimators': 25} |

## IV. METHODOLOGY

*Evaluation Matrices*

The dependent variable that we used in this study was binary, and it was recoded in Python in such a way that a minor injury would be represented by the number 0, while a serious injury would be represented by the number 1. To make classification models more easily interpretable, we assigned a positive value to serious injuries while assigning a negative value to less serious injuries. In the event of a collision, there are only four possible outcomes for the driver's injury situation, as detailed below:

**True positive (TP):** Prediction is positive, and injury is serious.
**True negative (TN):** Prediction is negative, and injury is non-serious.
**False positive (FP):** Prediction is positive, but injury is non-serious.
**False negative (FN):** Prediction is negative, but injury is serious.

Following are some evaluation matrices that we employed in order to evaluate the performance of models:

*Accuracy:* It is the proportion of subjects whose labels are accurate to the total number of subjects in the study. Accuracy is the one that comes to mind first.

$$Accuracy = \frac{TP + TN}{TP + TN + FP + FN}$$

*Precision:* Precision can be defined as the proportion of correctly positive values that a model assigns out of the total number of positive values it assigns.

$$\text{Precision} = \frac{TP}{TP + FP}$$

*Recall:* It is the proportion of drivers who are accurately identified as positive by a model compared to the total number of drivers who are at risk of serious injury.

$$\text{Recall} = \frac{TP}{TP + FN}$$

*F-1 score:* It is the harmonic mean, sometimes known as the average, of the recall and the precision.

$$\text{F1 score} = \frac{2\,(Recall \times Precision)}{Recall + Precision}$$

*ROC curve:* A receiver operating characteristic curve, often known as a ROC curve, is a graph that depicts the performance of a classification model across all classification levels. More commonly, a ROC curve is referred to as the acronym ROC. This curve depicts the relationship between two separate rates: the True Positive Rate (TPR) and the False Positive Rate (FPR). Both of these rates are denoted by their respective abbreviations. When the classification threshold is lowered, more things are tagged as positive, which increases the amount of both false positives and true positives. When the threshold is increased, a smaller percentage of the items are determined to be positive. A typical ROC curve is depicted in Figure 3.

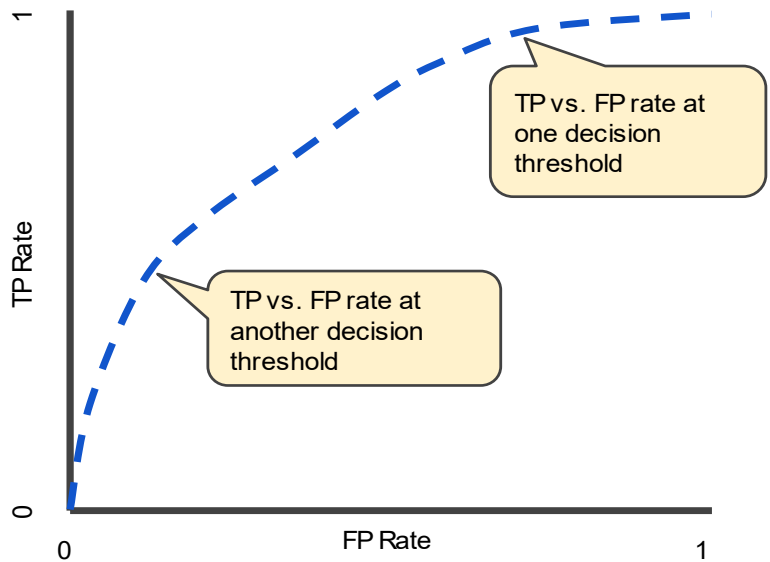

Fig. 3. Common ROC curve [source]

*Area under the ROC curve (AUC):* The area under the ROC curve from the origin (0,0) to the point of intersection (1,1) is what the AUC attempts to quantify. The area that is beneath a typical ROC curve is seen in figure 4.

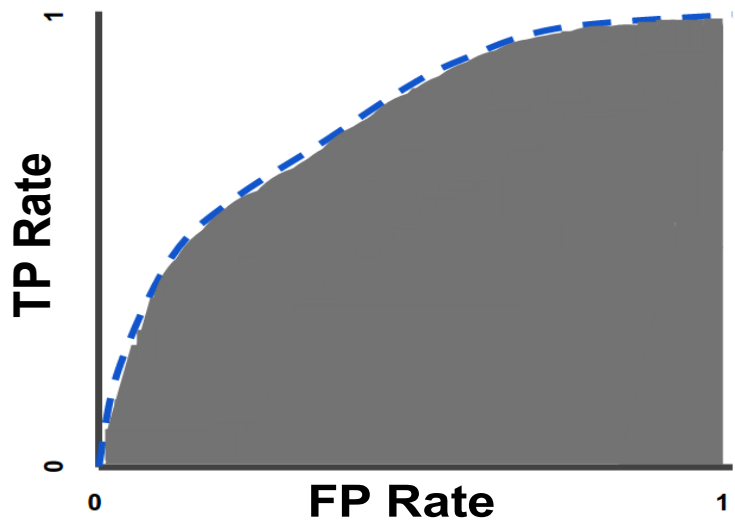

Fig. 4. The region under the ROC curve (also known as AUC) [source]

Our primary objective was to make an accurate prediction of the severity of the injuries sustained by drivers, and we deemed serious injuries to be more important than less serious ones. Precision takes into account both the true and false positive rates; nonetheless, if our model forecasts a minor injury as a significant injury, this wasn't a key concern for our research. We placed a greater emphasis on correctly forecasting a significant injury (abbreviated as TP) as opposed to incorrectly predicting a serious injury (abbreviated as FN). To solve this problem, we make advantage of the recall value. In order to evaluate the effectiveness of the various models, we compared their accuracy, recall, and AUC values.

## V. RESULTS

We utilized Python's 'scikit-learn' package in order to execute a number of different classification models on the dataset. We began by applying logistic regression to our dataset in order to analyze the features that were supplied. The greatest values for accuracy and area under the curve for logistic regression were 0.8938 and 0.4902, respectively.

We found that the decision tree gave us the 74% accuracy. The AUC was 0.5521, and the recall was 0.3194. When we used the Random Forest classifier, we were able to achieve an accuracy of 0.7574. The maximum recall value that Random Forest could achieve was 0.2986, which is virtually indistinguishable from the decision tree classifier. An AUC value of 0.5667 was obtained from Random Forest.

The level of accuracy that could be achieved with SVM was 75%, which is on par with what Random Forest achieved. Both the recall and AUC scores for SVM came in at 0.375 and 0.624 respectively. The highest accuracy achieved using Adaboost was 0.7588. Both the maximum recall and the AUC were equal to 0.3958 and 0.6420 respectively. XGBoost achieved 70% accuracy. Table IV presents the results of a comparison of the models for a variety of feature combinations.

*Baseline Methods:*

As a baseline method for this research, the SVM classifier was chosen. When contrasted with the outputs of the SVM classifier, the performances of the other models were evaluated.

TABLE IV. COMPARISON OF CLASSIFIER MODELS

| Model Name | Accuracy | Recall | AUC |
|---|---|---|---|
| Logistic Regression | 0.8938 | 0.0000 | 0.4902 |
| Decision tree | 0.7412 | 0.3194 | 0.5521 |
| Random forest | 0.7574 | 0.2986 | 0.5667 |
| SVM | 0.7500 | 0.3750 | 0.6240 |
| Adaboost | 0.7588 | 0.3958 | 0.6460 |
| XGBoost | 0.7080 | 0.3542 | 0.5718 |

Adaboost was successful in achieving recall and AUC values that were higher than those of other models. In spite of the fact that the logistic regression approach attained a maximum accuracy of 89%, we chose to employ the Adaboost outputs since our primary objective was to reliably forecast major accident injuries more so than other non-serious injuries. The ROC curves of six different classifiers are depicted in Figure 5. The Adaboost classifier outperforms its competitors.

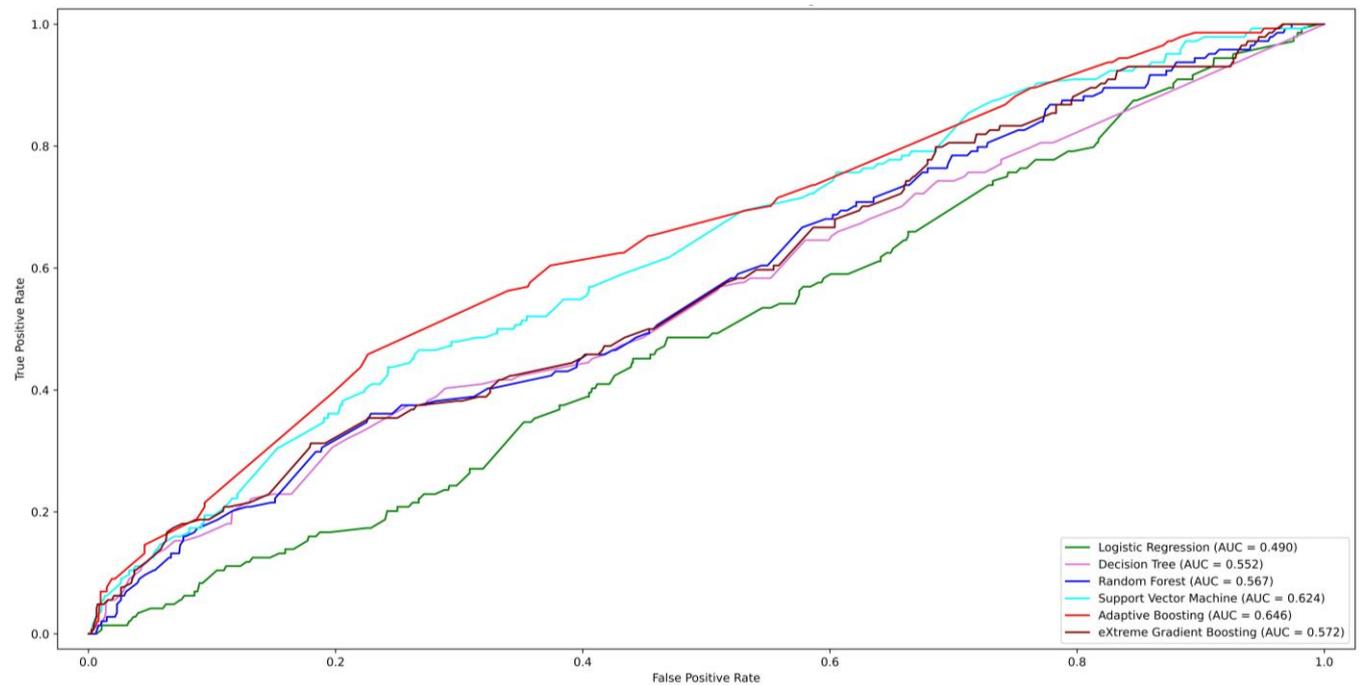

Fig. 5. ROC curves

*Shapley Additive exPlanations (SHAP)*

In the current investigation, the SHAP approach was utilized to rank the factors that were anticipated to play a role in the severity of driver injuries received as a result of collisions in order to establish their relative significance. By taking the absolute Shapley value of each feature and averaging it, (Fig. 6) shows how much each feature contributed to the overall accuracy of the forecast. The 'crash_type_Rear_End' has the most influence, as determined by the SHAP methodology, on whether or not a serious collision takes place. It was also observed that 'Area_type_Rural' was an important factor in predicting the severity of crash injuries.

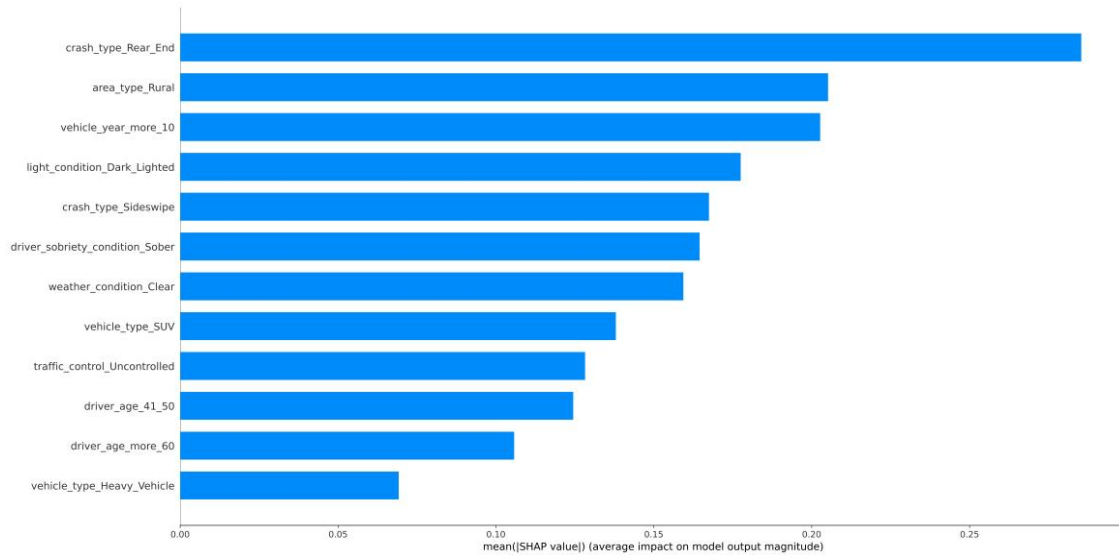

Fig. 6. Mean SHAP value of different features.

Fig. 6 illustrates the value of all features on a global scale which is one of its major drawbacks. Individual variables and the SHAP value associated with them were investigated further in order to ascertain their local magnitude and the contribution they made to the process of forecasting the severity of drivers' injuries sustained in crashes. The local explanation summary of all of the selected explanatory elements is depicted in figure 7. The color red implies that the corresponding variable has a higher value. In addition, SHAP values that are larger than zero (a higher value) indicate that they have a positive effect on the likelihood of inflicting a substantial injury. On the other hand, SHAP values that are less than zero (a lower value) indicate that they have a negative effect on the severity of the crash injury. According to the synopsis of the local explanation, accidents of the 'crash_type_Rear_End' variety are less likely to result in significant injuries. There is a correlation between the age of a vehicle and old vehicles have high likelihood of being involved in a serious collision than new vehicles.

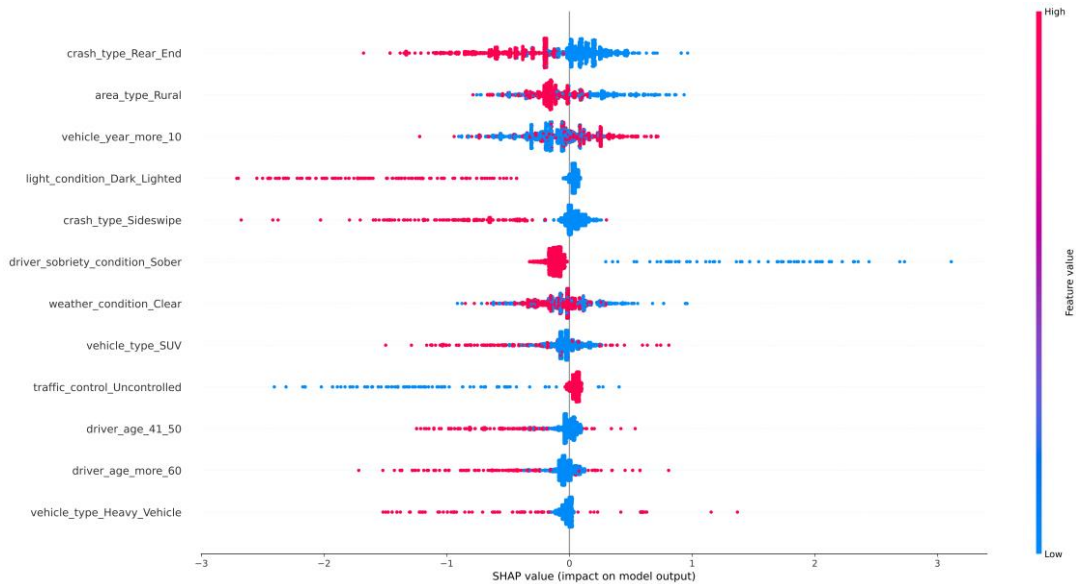

Fig. 7. Importance of features from SHAP value.

## VI. CONCLUSION AND DISCUSSIONS

In this study, a total of six different classification models were used to predict the severity of driver crash injuries. These models included logistic regression, a decision tree classifier, a random forest classifier, a support vector machine (SVM), and an adaptive boosting (Adaboost) classifier. Signal Four Analytics (S4A) was the company that supplied the dataset along with the necessary properties. The accuracy, recall, and area under the curve (AUC) values of each model's performance were compared to one another. Adaboost fared well in terms of recall and AUC value, however, logistic regression performed even better by having the highest accuracy value.

The dataset that was used for this research included a large amount of bias, with the bulk of the observations being information concerning less serious injuries. This limited the accuracy of the findings significantly. When using this type of data, the overall accuracy is often quite good as a result of the models' primary focus on attempting to forecast the major outcome, which is non-serious damage being caused to the participant. In other words, the models are trying to determine how likely it is that the participant will sustain a minor injury. Because our primary goal was to create a data-driven strategy for accurately projecting serious accident injuries over non-serious injuries, it would not be sufficient to simply evaluate whether or not the model was accurate. Rather, we needed to build a strategy for accurately forecasting significant crash injuries over non-serious injuries. As a direct consequence of this, we focused more on recall and AUC values than we did on accuracy. The ability of a model to recall desirable outcomes, which in our case are serious injuries sustained by a minority, is evaluated using the recall method. We select the model that has the best possible combination of accuracy, recall, and AUC value so that we may get the most accurate results from this study. The most successful model was Adaboost. In spite of the fact that the Adaboost classifier is not as accurate as logistic regression in predicting substantial crashes (i.e., the minor crash severity outcome), it is extremely accurate in predicting small crashes.

The contribution of each modeling element to the injury severity prediction of a crash was evaluated with the use of an XGBoost classifier that was trained with SHAP techniques. The 'crash_type_Rear_End' was the most significant variable that contributed to the severity of driver crash injuries, as stated by SHAP. In addition, it was discovered that older automobiles have a high likelihood of sustaining significant injuries in a collision.

There are some restrictions that apply to this study. We used the normal crash data that was reported. On the other side, predictive accident injury modeling has been shown to be beneficial. This is accomplished by combining skewed real-time historical collision data with data from traffic sensors, weather stations, and other data sources. In subsequent research, the prediction of the severity of a crash might be tested by using real-time data. There were six different classifications algorithms used in the study. Additionally, it has been suggested that more machine-learning strategies be applied to the same dataset in order to determine whether or not the recall score improves.